# MambaPlace: Text-to-Point-Cloud Cross-Modal Place Recognition with Attention Mamba Mechanisms

Tianyi Shang[1,2], Zhenyu Li[1,*], Pengjie Xu[3], Jinwei Qiao[1]

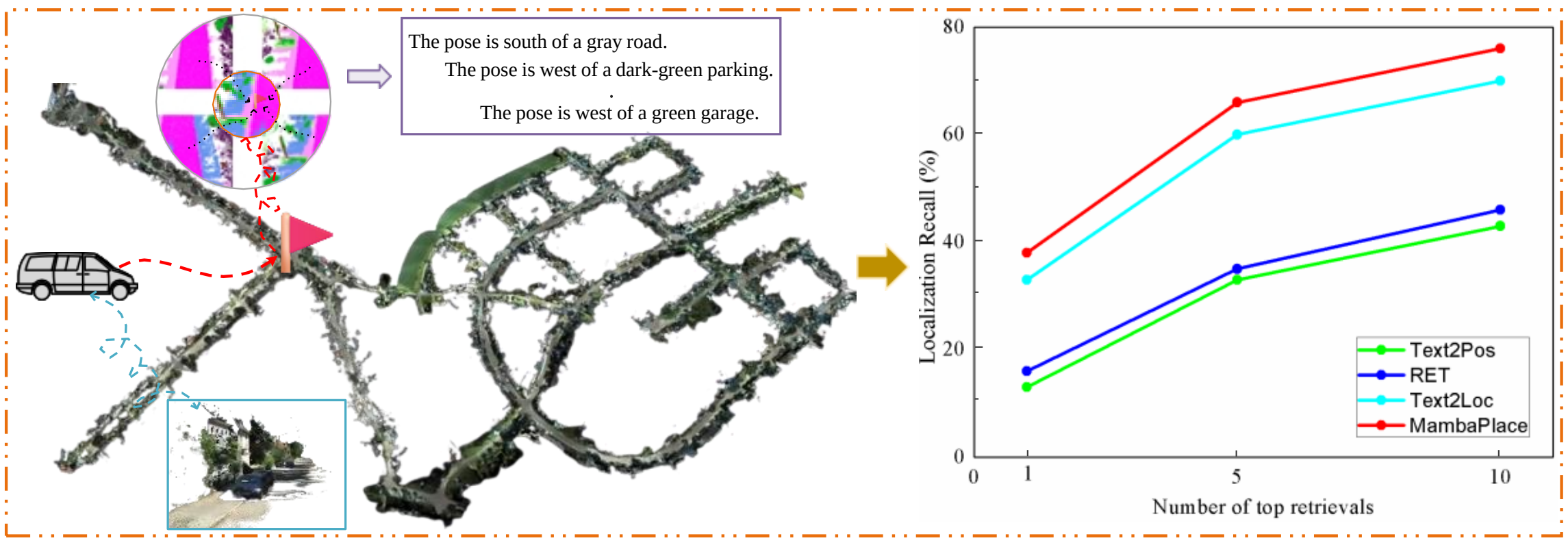

Fig. 1: Left: We introduce MambaPlace, a city-scale place localization solution that utilizes text descriptions. When given a point cloud representing the environment and a text query describing a place, MambaPlace identifies the most probable location of the specified place on the map. Right: The localization performance on the KITTI360Pose test set demonstrates that the proposed MambaPlace implementation consistently outperforms existing methods across all top search numbers. Notably, its performance in text localization surpasses all current SOTA results for queries within a 5-meter range.

***Abstract*— Vision-Language Place Recognition (VLPR) enhances robot localization performance by incorporating natural language descriptions from images. By utilizing language information, VLPR directs robot place matching, overcoming the constraint of solely depending on vision. However, general multimodal information integration methods are not well equipped to capture the dynamics of cross-modal interactions, especially in the presence of complex intra-modal and inter-modal correlations. To this end, this paper proposes a novel coarse- to-fine and end-to-end connected cross-modal place recogni- tion framework, called MambaPlace. In the coarse-localization stage, the text description and 3D point cloud are encoded by the pre-trained T5 and instance encoder, respectively. They are then processed using Text-Attention Mamba (TAM) and Point Cloud Multi-Strategy Scanning Mamba (MSSM), with the latter mimicking the eye's focusing mechanism, for data enhancement and alignment. In the subsequent fine-localization stage, the features of the text description and 3D point cloud are cross-modally fused and further enhanced through Cascaded Cross-Attention Mamba (CCAM). Finally, we predict the positional offset from the fused text-point cloud features, achieving the most accurate localization. Extensive experiments show that MambaPlace achieves improved localization accuracy on the KITTI360Pose dataset compared to the state-of-the-art methods. Specifically, as shown in Fig. 1, when $\epsilon$<5, MambaPlace achieves 5% higher test accuracy compared to the existing state-of-the-art. Our code is available at https://github.com/nuozimiaowu/MambaPlace/tree/main.**

## I. INTRODUCTION

In future smart cities, autonomous vehicles, drones, and intelligent logistics systems will need to accurately localize based on human language descriptions before effective path planning can occur. Traditional unimodal visual place recognition (VPR) methods rely on cameras or radar to extract features from 2D images or point clouds and subsequently retrieve corresponding locations from a database. However, these methods suffer from low efficiency in human-computer interaction and lack precision under conditions of seasonal changes and variations in viewpoint. In contrast, the text- to-point-cloud localization approach enables accurate localization without requiring the user to be in proximity to the location.

Text-to-point-cloud localization encounters several challenges: 1) Ambiguous descriptions may correspond to multiple potential regions within the point cloud map, and 2) language descriptions of different positions within the same region can be very similar, making precise location regression a significant challenge. To address these issues, the pioneering work Text2Pos generated multiple descriptions for various spatial positions based on the KITTI360 dataset, thereby creating the foundational KITTI360Pose dataset. Subsequently, they proposed the first two-stage text-to-point-cloud localization framework: In the coarse phase, large-scale point clouds are segmented into patches and aligned with text; in the fine phase, text and point cloud fusion is employed to achieve precise localization within each candidate region. However, Text2Pos focuses primarily on descriptions within patches and overlooks the global spatial relationships between language and point clouds. To address this limitation, Wang et al. introduced the Relation-Enhanced Transformer (RET) to establish relationships between point clouds and text [1], utilizing cross-attention in the fine phase to enhance multimodal fusion. Recently, Text2loc [2] introduced a contrastive learning mechanism in the coarse phase, significantly improving performance.

*This work was supported by the Natural Science Foundation of Shandong Province (ZR2024QF284), the Opening Foundation of Key Laboratory of Intelligent Robot (HBIR202301), the Open Project of Fujian Key Laboratory of Spatial Information Perception and Intelligent Processing (FKLSIPIP1027). *(Corresponding authors: Zhenyu Li)*

[1]Tianyi Shang, Zhenyu Li, and Jinwei Qiao are with the School of Mechanical Engineering, Qilu University of Technology (Shandong Academy of Sciences), Jinan 250353, China (832201319@fzu.edu.cn; lizhenyu@qlu.edu.cn)

[2]Tianyi Shang is also with the Department of Electronic and Information Engineering, Fuzhou University, Fuzhou 350100, China (832201319@fzu.edu.cn)

[3]Pengjie Xu is with the School of Mechanical Engineering, Shanghai Jiao Tong University, Shanghai 200030, China (xupengjie194105@sjtu.edu.cn)

However, previous methodologies continue to exhibit unresolved challenges. While language information is represented using the advanced T5 model, point cloud feature extraction relies on a simple encoder. This approach fails to adequately capture the intricate features of complex point clouds, especially position relevance between objects, resulting in an imbalance between text and point cloud.

In response to these limitations, we examine the potential of the Mamba architecture as a viable solution. Mamba is based on the state space model (SSM), which has achieved linear complexity and long-range context learning abilities in sequence data. Leveraging these characteristics of the Mamba framework, we propose MambaPlace, a novel framework for VLPR. Our main contributions are as follows.

- A Multi-Strategy Scanning Mamba (MSSM), utilizing pure selective SSM with various scanning strategies during point cloud encoding, mimicking the dynamic focusing mechanism of the retina and peripheral perception, thereby enhancing the relative positional relationships between objects in point clouds.
- A Text-Attention Mamba (TAM) utilizes Mamba's selective mechanism and linear processing capabilities to strengthen the semantic associations between directional and target keywords while enhancing sensitivity to directional keywords.
- A Cascaded Cross-Attention Mamba (CCAM), which can be interpreted as a specialized SSM with hidden states influenced by cross-modal attention mechanisms, is proposed to facilitate multi-scale multimodal feature fusion, thereby accurately predicting localization offsets and enhancing localization accuracy.

## II. RELATED WORK

### A. Place Recognition

*1) Visual Place Recognition:* Visual Place Recognition (VPR) involves accurately identifying a location by retrieving relevant information from a large-scale database. Traditional 2D VPR techniques employ various aggregation methods, such as Vector of Locally Aggregated Descriptors (VLAD) [3] and Generalized Mean (GeM), to extract feature vectors from images and then perform matching between 2D features. However, these methods often demonstrate suboptimal performance under conditions such as changes in viewpoint and seasonal variations. To address these challenges, Izquierdo et al. [4] explored the bidirectional relationship between clusters and features, Lu et al. [5] implemented feature alignment at the image scale, and EffoVPR [6] utilized attention maps from intermediate layers of Vision Transformers (ViT) for local matching, thereby enhancing performance across varying perspectives. Li et al. [7] proposed a novel adaptive soft-hard label teaching feature-level knowledge distillation learning framework for all-day mobile robot VPR tasks.

*2) LiDAR Place Recognition:* Recent studies increasingly focus on exploring point cloud-based LiDAR Place Recognition (LPR). To aggregate sparse and unordered point cloud features, previous methods employed handcrafted feature extraction techniques for point cloud representation, which were subsequently aggregated using VLAD to generate abstract descriptors. Current research is shifting towards end-to-end approaches. For example, ComPoint [8] mapped point cloud information to the complex domain, HiBi-GCN [9] introduced a hierarchical bidirectional graph convolutional network, and Xia et al. [10] proposed a two-stage cross-attention transformer (CASSPR) that combines point clouds and voxels, demonstrating excellent performance in point cloud representation.

*3) Cross-modal Place Recognition:* Recently, cross-modal VPR involving text and point clouds has garnered significant attention. Text2pos [11] pioneered this field by proposing a two-stage coarse-to-fine framework, where text-to-submap instance matching is performed in the coarse stage, followed by instance-level offset regression in the fine stage to achieve precise localization from text-to-point clouds. RET enhanced semantic representation and modality fusion through attention mechanisms. Text2Loc [2] significantly improved performance in the cross-modal domain by utilizing contrastive learning and introducing a novel non-matching localization method in the fine stage to reduce computational overhead.

### B. State Space Model

To address the limitations of CNNs in understanding large-scale scene representations and the quadratic computational cost associated with the self-attention mechanism as the number of tokens increases, the State Space Model (SSM) has emerged as a viable solution. SSM effectively captures implicit relationships between inputs and outputs and is widely utilized in fields such as automatic control theory and reinforcement learning. Recently, inspired by classical time series models such as RNNs and LSTMs, structured SSMs have been integrated into neural networks [12], [13]. Gu et al. [14] introduced the Mamba module, a selective SSM that enhances the time-invariance of the SSM by selectively discarding information. Unlike Transformers, the computational cost of the Mamba model increases linearly with the sequence length, making it efficient in managing long sequences and demonstrating significant potential in various downstream tasks.

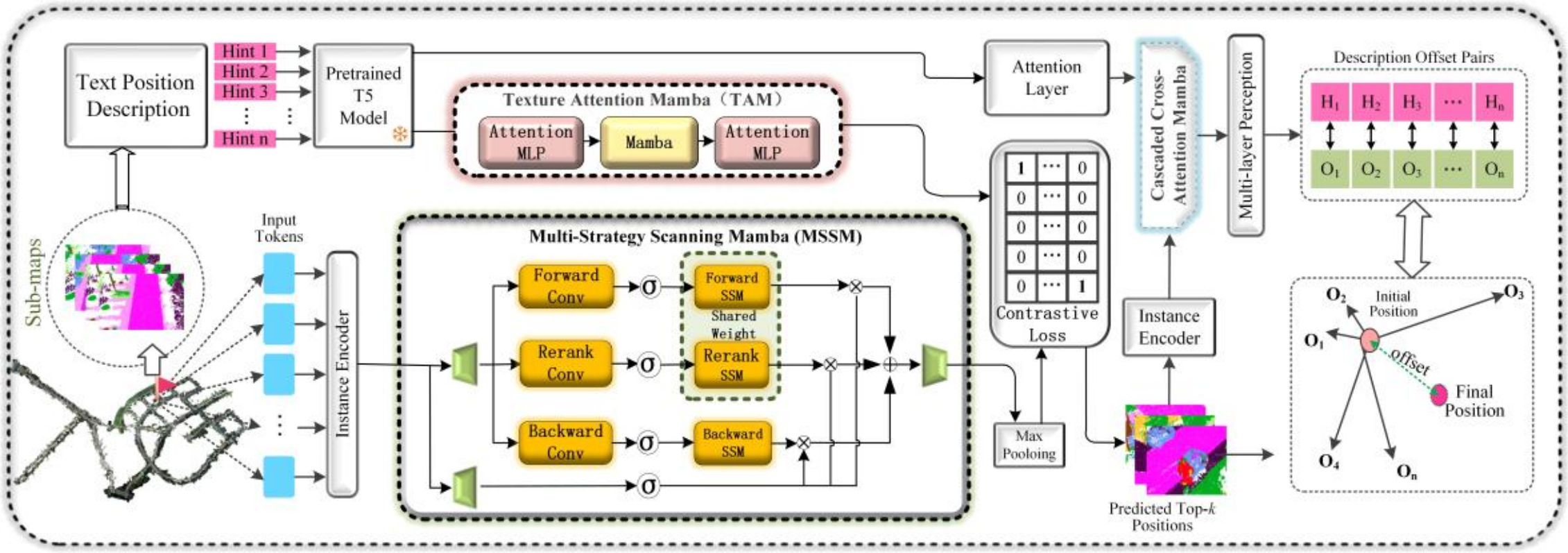

Fig. 2. The proposed MambaPlace architecture comprises two sequential modules: global coarse localization and fine localization. In the coarse localization stage, the text description and 3D point cloud are encoded using the pretrained T5 model and an instance encoder, respectively. These encodings are then processed with TAM and MSSM for location enhancement and alignment. In the subsequent fine localization stage, the features from the text description and 3D point cloud are cross-modally fused using CCAM. Finally, the relative positional offset of each point cloud submap's coordinates to the center point is obtained to adjust the target location, thereby improving localization accuracy.

Mamba has been extensively utilized in various computer vision tasks. Vision Mamba enhanced the original Mamba's Single Shot Detector (SSD) module, enabling it to process 2D data and thereby initiating the application of Mamba in the field of computer vision. Combined Vision Transformers (ViT) and Mamba, Zhu et al. [15] introduced the pure SSM-based Vim method, which demonstrated exceptional performance across several downstream tasks. Currently, Mamba has been widely used in object detection [16], semantic segmentation [17], and image retrieval [18]. The effectiveness of SSM models in point cloud processing was illustrated by PoinTramba [19] and PointMamba [20]. Dimba [21], a novel text-to-image diffusion model, showcased Mamba's capabilities in cross-modal processing. Mamba2 [22] further improved Mamba by integrating attention mechanisms and enhancing parallel processing.

Our proposed MambaPlace framework pioneers the integration of State Space Models into cross-modal Place Recognition tasks. Through our innovatively designed MSSM and TAM, they enhance the representation of key features in point clouds and textual descriptions, with particular emphasis on spatial-positional characteristics. Furthermore, the CCAM facilitates comprehensive fusion between point cloud and textual modalities. Extensive experiments demonstrate that MambaPlace significantly outperforms existing state-of-the-art methods across various benchmarks.

## III. METHODOLOGY

MambaPlace represents large-scale 3D maps as $M_{ref} = \{m_i : i = 1, ..., M\}$, a collection of cubic submaps $m_i$, where each submap contains a set of 3D object instances $I_{i,j}$. Let T represent the description of the query text, which consists of a set of hints $\{\vec{h}_k\}_{k=1}^{h}$, each hint describing the spatial relationship between the target locations of an object instance. As shown in Fig. 2, to effectively process these multimodal data, we divide our framework into two consecutive end-to-end stages: global coarse localization and fine localization.

The global coarse stage aims to train a function F that encodes $T_i$ and $m_i$ into a unified embedding space, using contrastive learning as the loss function, where the semantic similarity between matching pairs is maximized.

The fine stage aims to directly regress the object's location directly coordinates $(x, y)$ in the scene coordinate system. Given the text description $T_i$ and submaps $m_i$ retrieved from the coarse stage, we formulate the localization objective as:

$$\min E \left\| (x, y) - \phi\left(T_i, \operatorname*{argmin}_{m_i \in M_{\mathrm{ref}}} \{ d(F(T_i), F(m_i)) \}\right) \right\|^2 \tag{1}$$

Where $d(\cdot)$ is the Euclidean distance between the text descriptor and point cloud descriptor, $\phi$ is an MLP network to predict coordinates.

### A. Preliminaries: SSM and Mamba

*1) State Space Model:* The State Space Model (SSM) is a linear time-invariant system where the current state exclusively determines input at each timestep, enabling efficient sequential processing of information. Given an input sequence $u(t) \in \mathbb{R}^L$, it first updates the current hidden state $x(t) \in \mathbb{R}^N$ using the current input $u(t) \in \mathbb{R}^L$ and previous hidden state, then generates an output $y(t) \in \mathbb{R}^L$ based on the updated current hidden state. The state space dynamics are expressed as follows:

$$\begin{aligned} x'(t) &= Ax(t) + Bu(t) \\ y(t) &= Cx(t) + Du(t) \end{aligned} \tag{2}$$

Where $A \in \mathbb{R}^{N \times N}$ is the state matrix, $B \in \mathbb{R}^{N \times L}$ is input matrices, $C \in \mathbb{R}^{L \times N}$ is output matrices and $D \in \mathbb{R}^{L \times L}$ is set to zero to eliminate direct feed through from input to output.

*2) Discretization and Parallelization:* For a discrete sequence $(u_0, u_1, ...) \in \mathbb{R}^L$, state space model can be discretized into the following form:

$$\begin{aligned} x_k &= \bar{A}x_{k-1} + \bar{B}u_k \\ y_k &= \bar{C}x_k \end{aligned} \tag{3}$$

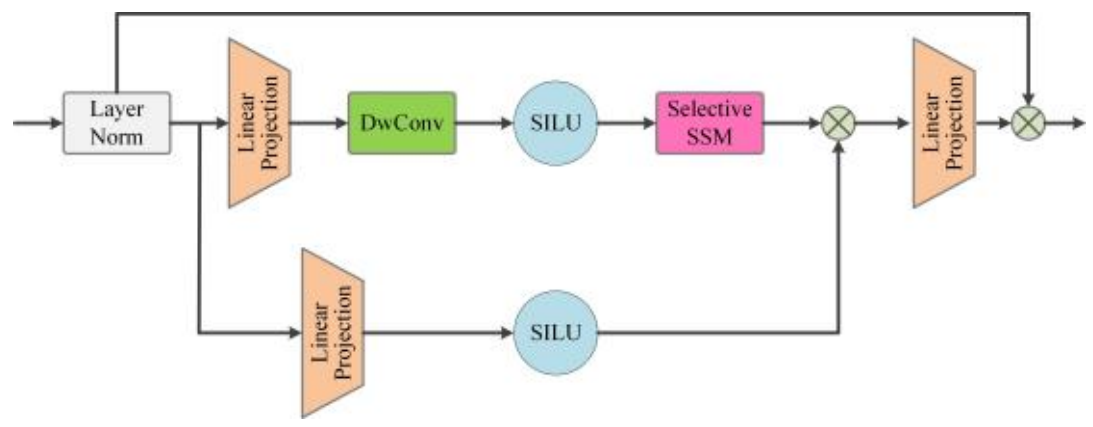

Fig. 3. Basic Architecture of Mamba: A Selective State Space Model for Sequential Processing.

The continuous-time parameters A and B can be transformed into $\bar{A}$ and $\bar{B}$ through numerical approximation methods such as zero-order hold (ZOH).

$$
\begin{aligned}
\bar{A} &= \exp(\Delta A) \\
\bar{B} &= (\Delta A)^{-1}(\exp(\Delta A) - I) \cdot \Delta B
\end{aligned}
\tag{4}
$$

For an input sequence of length $M$, the discrete SSM computation can be parallelized through a global convolution with kernel $\bar{K}$.

$$
\bar{K} = (C\bar{B}, C\bar{A}\bar{B}, \dots, C\bar{A}^{M-1}\bar{B}), \quad y = x * \bar{K} \tag{5}
$$

*3) Selective State Space Model:* Traditional State Space Models (SSMs) employ linear processing uniformly across all inputs, failing to capture the inherent heterogeneity in input importance. As shown in Fig. 3, to overcome this fundamental limitation, Mamba introduces the S6 algorithm, which transforms conventional SSMs into time-varying systems by making parameters $\Delta$, B, and C input-dependent. Specifically, these parameters undergo dynamic adjustment through learnable fully connected layers conditioned on the input, enabling the model to selectively attend to salient information. The S6 algorithm preserves the computational efficiency inherent to SSMs while substantially enhancing the model's capacity for long sequence processing through its selective information processing mechanism.

### B. Global Coarse Localization

Coarse place retrieval comprises two parallel branches: text encoding and point cloud encoding. To reinforce the cross-modal feature alignment between text and point cloud modalities for identical objects and enhance the feature representation, we reconstruct the encoders in both branches by building the attention Mamba mechanism.

*1) Text Encoding Branch:* Considering that our textual descriptions follow a 'location + object' structure, and objects frequently recur in scenes, spatial terms play a crucial role in the text localization task. Based on this observation, we propose the TAM module, which leverages the selective mechanism of Mamba to specifically enhance the feature representation of spatial descriptions, thereby improving localization accuracy.

The textual information is initially processed using a large pre-trained T5 model, which has fixed parameters to fully embed linguistic characteristics. The embedded text then enters our custom-designed TAM module. As shown in Fig. 2, the TAM module employs a stacked architecture consisting of attention and Mamba components. Initially, a word-level Transformer captures contextual relationships between words, followed by a multi-layer perceptron (MLP) that generates comprehensive sentence-level descriptors. These descriptors are then processed by the Mamba block, where its selective SSM mechanism effectively filters out irrelevant information while preserving crucial spatial orientations and object-specific terms. Notably, Mamba's linear processing capability enhances the model's discrimination of directional terms across multiple descriptions of the same location. The framework culminates with additional transformer and MLP layers that aggregate multiple sentence descriptions into a global semantic descriptor, enriched with spatial positional information derived from key terms.

*2) Point Cloud Encoding Branch:* Compared to textual modalities, point cloud data exhibits more complex spatial relationships. To enhance the relative positional relationships between different objects, we propose a novel Multi-Strategy Scanning Mamba (MSSM). Built upon the Mamba architecture, MSSM draws inspiration from the human visual system, specifically mimicking the dynamic focusing mechanism of the retinal fovea and the broad perceptual capabilities of peripheral vision. As shown in Fig. 2, our point cloud encoding branch consists of two main components: a PointNet++ backbone for initial feature extraction, followed by our proposed MSSM for enhanced point cloud spatial relationship modeling. The MSSM incorporates parallel scanning mechanisms: *dual-path forward scanning* that simulates foveal fixation and *reverse scanning* that emulates broad peripheral vision.

- *dual-path forward scanning:* For an input sequence $X = \{x_1, x_2, \dots, x_n\}$, where $n$ denotes the sequence length, we implement two parallel inputs:

$$
\begin{aligned}
f(X) &= \{x_1, x_2, x_3, \dots, x_n\} \\
g(X) &= \{x_{2i}, x_{2i-1} \mid i \in \{1, 2, \dots, \lfloor \tfrac{n}{2} \rfloor\}\}
\end{aligned}
\tag{6}
$$

  The dual-path output is formulated as:

$$
\mathrm{MSSM}_{\mathrm{loc}}(X) = \mathrm{Concat}(\mathrm{Mam}[f(X)], \mathrm{Mam}[g(X)]) \tag{7}
$$

  As shown in Fig. 4(left), $f(X)$ and $g(X)$ facilitate the reorganization of adjacent processing units into new fundamental computational blocks. Each locally restructured unit incorporates cyclic connections while preserving the global scanning properties of the system. This mechanism simulates microsaccades around the fixation point in human vision, enabling reciprocating attention computations between adjacent features.
- *reverse scanning:* To complement the global feature capturing, we incorporate a parallel reverse scanning mechanism:

$$
\mathrm{MSSM}(X) = \mathrm{Concat}(\mathrm{Mam}[f(X)_{\mathrm{rev}}], MSSM_{\mathrm{loc}}(X)) \tag{8}
$$

Where Mam $(\cdot)$ is the vector output of the Mamaba model, $\mathrm{MSSM}_{loc}(\,)$ is the vector output of the multi-strategy scanning Mamba model.

This mechanism emulates the broad perceptual capabilities of retinal peripheral vision, capturing long-range position

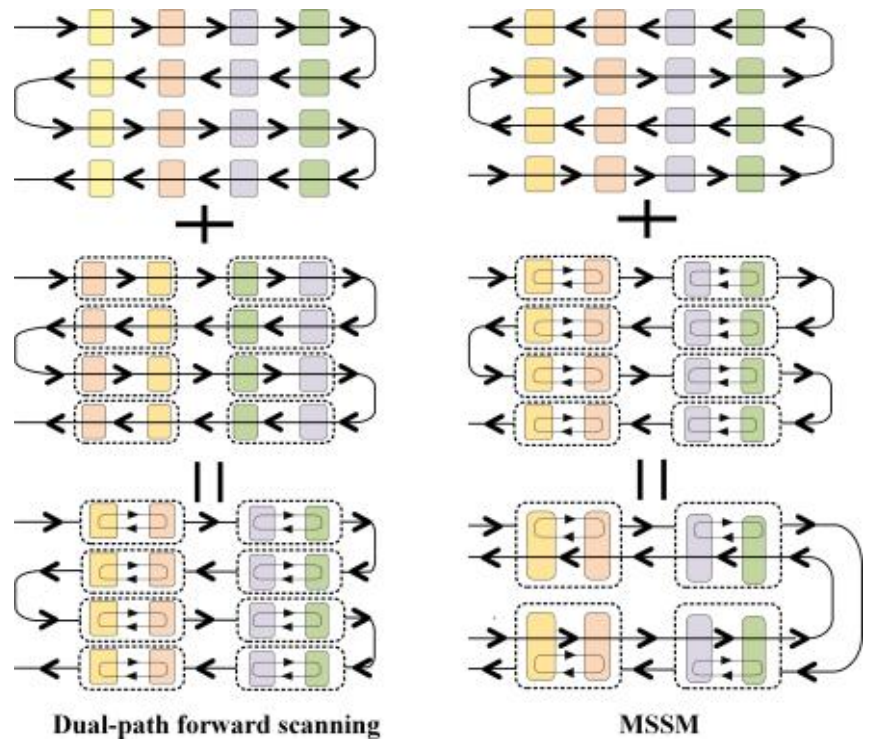

Fig. 4. Left: Illustration of MSSM's dual-path forward scanning mechanism, showing parallel processing of original sequence $f(X)$ and reorganized sequence $g(X)$, with cyclic connections simulating foveal microsaccades. Right: The complete MSSM architecture incorporates a reverse scanning mechanism on top of dual-path forward scanning to emulate human peripheral vision, enabling comprehensive capture of both local details and global spatial relationships in point cloud data.

dependencies in the point cloud data.

As illustrated in Fig. 4(right), the complete MSSM architecture integrates both local and global spatial relationships through its multi-directional scanning mechanisms. The dual-path forward scanning captures adjacent object location relationships by simulating microsaccades, while the reverse scanning captures broad contextual information through peripheral vision simulation. This comprehensive design not only enhances the positional relationships between different objects in the point cloud but also facilitates the projection of point cloud features into a semantic space that aligns with text encodings, enabling effective cross-modal learning.

### C. Fine-grained Location

In the fine localization stage, MambaPlace has identified multiple potential corresponding point cloud regions for each text description. The objective of this stage is to regress the position of each pair of language and point cloud regions, ultimately obtaining the precise coordinates that correspond to each text description. To address this challenge, we present a novel cascaded cross-attention Mamba model (CCAM, as shown in Fig. 5), which can be interpreted as a specialized variant of SSM and effectively integrates information from both text and point clouds.

First, the point cloud features and text features are processed through the MHA mechanism. The point cloud acts as the query Q, while the texture features serve as the key K and value V:

$$Z_1 = \text{Add\&Norm}(Q, \text{MHA}(Q, K, V)) \tag{9}$$

Then, Z is processed through the FFN:

$$Z_1' = \text{Add\&Norm}(Z_1, \text{FFN}(Z_1)) \tag{10}$$

The Mamba block further processes the features. Here, it is assumed that the Mamba Block generates new queries, keys, and values:

$$Q', K', V' = \text{Mam}(Z_1') \tag{11}$$

Then, these are further processed by the cross-attention module:

$$H_1 = \text{CrossAtten}(Q', K', V') \tag{12}$$

The combination of Mamba block and cross-attention module represents the cascade processing of cross-modal deep feature fusion:

$$Q_i', K_i', V_i' = \text{Mam}_i(H_1) \tag{13}$$

Here, i represents the processing at each layer, with the final output being final $H_{final}$. Combining the above steps, the three key formulas can be summarized as follows:

$$H_{final} = \text{CrossAtten}(\text{Mam}_n(\text{CrossAtten}(\text{Mam}_1(H_1)), \ldots)) \tag{14}$$

Where n is the number of Mamba blocks.

CCAM can be interpreted as a specialized State Space Model (SSM), where cross-modal attention operations serve as interaction mechanisms between the hidden states of point cloud and textual modalities. During the sequential processing of both modalities, dynamic interactions between hidden states enable deep inter-modal information exchange. The multi-layer cascaded structure progressively optimizes these hidden states, ultimately achieving comprehensively fused cross-modal feature representations.

The fine localization network does not include a text instance matching process, which simplifies and accelerates model training. The primary objective of the fine stage is to minimize the discrepancy between the predicted location and the actual location of the object. Specifically, the fine stage only utilizes the mean squared error loss to train the translation regressor.

$$L(C_{gt}, C_{pred}) = \|C_{gt} - C_{pred}\|_2 \tag{15}$$

Where $C_{pred} = (x, y)$ represents the predicted target coordinates, and $C_{gt}$ denotes the ground truth coordinates.

## IV. EXPERIMENTS

### A. Dataset

The KITTI360Pose dataset encompasses a total area of 15.51 square kilometers across nine urban regions, consisting of 43,381 point cloud-text pairs. In our experiments, we selected five regions (11.59 square kilometers) for the training set, one region for the validation set, and the remaining three regions for the test set. Each $30 \times 30$ meter point cloud region was treated as the smallest unit, with a stride of 10 meters between each unit.

### B. Implementation Details

In the coarse localization stage, we used a fixed learning rate of 5e-4 for position regression, training for 20 epochs. In the fine localization stage, we applied a fixed learning rate of 3e-4 for position refinement, training for 35 epochs. To comprehensively evaluate the experimental results, we

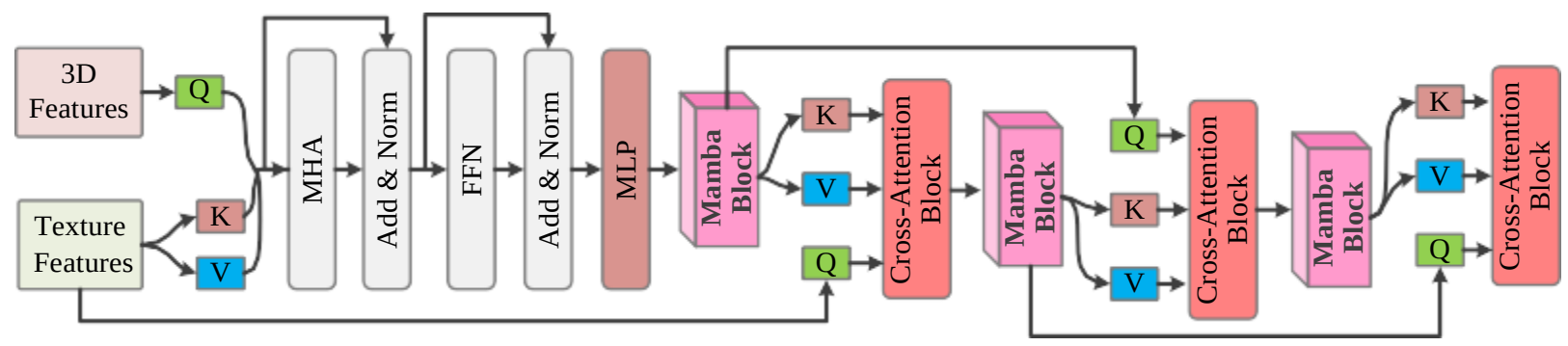

Fig. 5. The Cascaded Cross-Attention Mamba (CCAM) achieves comprehensive fusion between point cloud and textual representations through cross-branch information querying, utilizing cross-modal attention mechanisms while employing Mamba for linear processing. It can be interpreted as a specialized variant of SSM, where cross-modal attention enables parallel interactions between hidden states.

TABLE I

A COMPREHENSIVE COMPARISON BETWEEN MAMBAPLACE AND EXISTING SOTA METHODS ON THE KITTI360POSE DATASET.

| Methods | L calization Recall ($\epsilon$<5/10/15$m$) ↑ | | | | | |
|---|---|---|---|---|---|---|
| | Validation Set | | | Test Set | | |
| | k = 1 | k = 5 | k = 10 | k = 1 | k = 5 | k = 10 |
| NetVLAD [3] | 0.18/0.33/0.43 | 0.29/0.50/0.61 | 0.34/0.59/0.69 | — | — | — |
| PointNetVLAD [24] | 0.21/0.28/0.30 | 0.44/0.58/0.61 | 0.54/0.71/0.74 | 0.13/0.17/0.18 | 0.28/0.37/0.39 | 0.28/0.37/0.39 |
| Text2Pos [11] | 0.14/0.25/0.31 | 0.36/0.55/0.61 | 0.48/0.68/0.74 | 0.13/0.20/0.30 | 0.33/0.42/0.49 | 0.43/0.61/0.65 |
| RET [23] | 0.19/0.30/0.37 | 0.44/0.62/0.67 | 0.52/0.72/0.78 | 0.16/0.25/0.29 | 0.35/0.51/0.56 | 0.46/0.65/0.71 |
| Text2Loc [2] | 0.37/0.57/0.63 | 0.68/0.85/0.87 | 0.77/0.91/0.93 | 0.33/0.48/0.52 | 0.60/0.75/0.78 | 0.70/0.84/0.86 |
| MambaPlace (Our) | **0.45/0.62/0.68** | **0.75/0.89/0.90** | **0.83/0.94/0.95** | **0.38/0.52/0.55** | **0.66/0.79/0.81** | **0.76/0.87/0.89** |

TABLE II

ACCURACY OF MAMBAPLACE AND SOTA METHODS IN THE GLOBAL PLACE RECOGNITION STAGE ON THE KITTI360POSE DATASET.

| Methods | Submap Retrieval Recall ↑ | | | | | |
|---|---|---|---|---|---|---|
| | Validation Set | | | Test Set | | |
| | k = 1 | k = 3 | k = 5 | k = 1 | k = 3 | k = 5 |
| Text2Pos [10] | 0.14 | 0.28 | 0.37 | 0.12 | 0.25 | 0.33 |
| RET [22] | 0.18 | 0.34 | 0.44 | - | - | - |
| Text2Loc [2] | 0.31 | 0.54 | 0.64 | 0.28 | 0.49 | 0.58 |
| MambaPlace (Our) | **0.35** | **0.61** | **0.72** | **0.31** | **0.53** | **0.62** |

TABLE III

PERFORMANCE COMPARISON OF THE FINE-STAGE MODELS ON THE KITTI360POSE DATASET. NORMALIZED EUCLIDEAN DISTANCE IS ADOPTED AS THE METRIC. ALL METHODS DIRECTLY TAKE RAW POINT CLOUD AS INPUT.

| Methods | Validation Error ↓ | Test Error ↓ |
|---|---|---|
| Text2Pos | 0.120 | 0.121 |
| Text2Loc | 0.091 | 0.090 |
| MambaPlace (Our) | **0.086** | **0.084** |

considered different query boundaries $\epsilon(5, 10, 15)$. If the positional error fell within the query boundary, we classified it as a positive sample. The entire experiment was conducted using PyTorch on an Ubuntu 20.04 operating system, supported by a hardware platform featuring a 96-core AMD CPU and a 24GB RTX 4090 GPU.

### C. Performance Comparison with SOTA Methods

*1) The Overall Performance of the Entire Model:* We compared MambaPlace with previous SOTA methods on the KITTI360Pose dataset. We considered different query boundaries ($\epsilon = 5, 10, 15$) and top-k values ($k = 1, 5, 10$) to ensure a comprehensive evaluation. The results indicated that our method consistently and significantly outperformed the previous SOTA method under various conditions, as illustrated in Table I. Specifically, MambaPlace achieved top-1 recall rates of 0.45 and 0.38 on the validation and test sets, respectively, when $\epsilon < 5$m, representing improvements of **8%** and **5%** over Text2Loc. The top-1 recall rate at $\epsilon < 5$m was a critical metric for assessing localization accuracy, and MambaPlace's performance in this scenario underscored its superior ability to accurately localize compared to other methods.

*2) Performance Evaluation Based on Coarse Stage and Fine Stage:* MambaPlace first underwent the global place recognition stage. As shown in Table II, we compared our model with three other state-of-the-art methods, evaluating the top-k (k=1, 3, 5) accuracy on both the validation and test sets. The results indicated that our model achieved accuracies of 0.35, 0.61, and 0.72 on the validation set, and 0.31, 0.53, and 0.62 on the test set, representing improvements of **4%**, **7%**, and **8%**, and **3%**, **4%**, and **4%**, respectively, over the previous best method. These findings demonstrated that the enhancements we introduced for point cloud and semantic information were highly effective, enabling our method to significantly outperform previous SOTA methods in the global place recognition stage, thereby improving the overall accuracy of the framework. Furthermore, we recorded the entire training process on the training set, validation set, and test set, comparing the proposed MambaPlace with the best method, Text2Loc, as illustrated in Fig. 6.

In addition, we further verified the predicted offset error during the fine localization stage, as shown in Table III. It was evident that MambaPlace achieved a smaller deviation, demonstrating the superior effectiveness of our proposed CCAM.

### D. Ablation Study

*1) Ablation Study with Global Place Recognition:* We sequentially removed the T5 model, contrastive loss, MSSM, TAM, and T $AM_{mamba}$ for ablation experiments. The results in Table IV demonstrated that each module contributed to

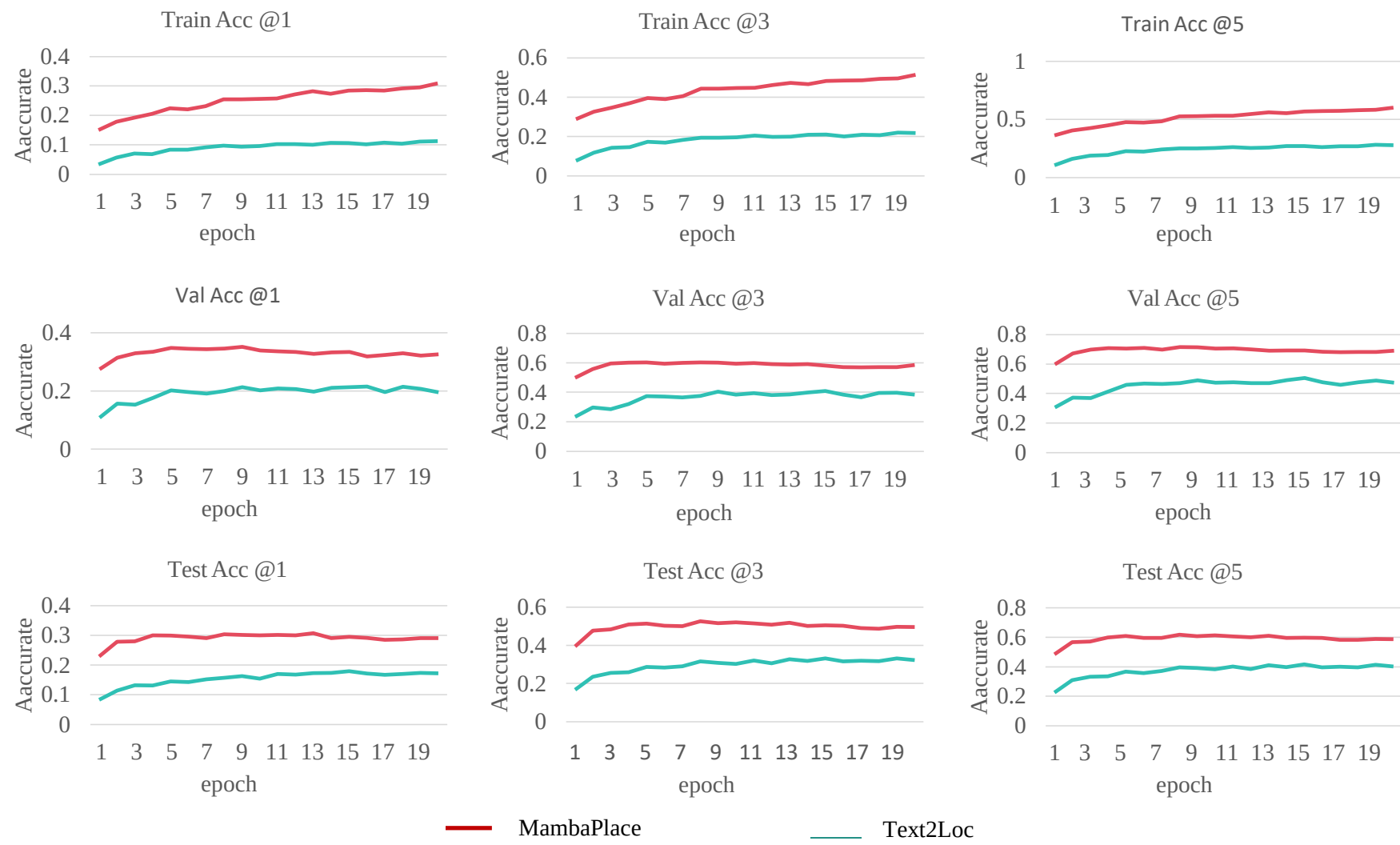

Fig. 6. Training results comparison between the proposed MambaPlace and Text2Loc based on the training, validation, and test sets, respectively.

TABLE IV

ABLATION STUDIES CONDUCTED ON THE KITTI360POSE DATASET DURING GLOBAL PLACE RECOGNITION REVEAL SEVERAL MODIFICATIONS. SPECIFICALLY, "W/O CL" DENOTES THE SUBSTITUTION OF THE CONTRASTIVE LOSS WITH A PAIRWISE LOSS, WHILE "W/O T5" INDICATES THE EXCLUSION OF THE PRE-TRAINED T5 MODEL AND "W/O TAM MAMBA" REPRESENTS THE REMOVAL OF THE MAMBA LAYER FROM THE TAM MODULE.

| Methods | Submap Retrieval Recall ↑ | | | | | |
|---|---|---|---|---|---|---|
| | Validation Set | | | Test Set | | |
| | K=1 | K=3 | K=5 | K=1 | K=3 | K=5 |
| W/o CL | 0.277 | 0.515 | 0.626 | 0.235 | 0.420 | 0.504 |
| w/o T5 | 0.312 | 0.550 | 0.669 | 0.274 | 0.456 | 0.577 |
| w/o MSSM | 0.213 | 0.419 | 0.519 | 0.186 | 0.341 | 0.425 |
| w/o TAM | 0.272 | 0.504 | 0.607 | 0.232 | 0.421 | 0.510 |
| w/o TAM_mamba | 0.348 | 0.601 | 0.711 | 0.321 | 0.505 | 0.604 |
| MambaPlace (Our) | **0.352** | **0.611** | **0.721** | **0.349** | **0.530** | **0.621** |

TABLE V

ABLATION STUDIES CONDUCTED ON MSSM REVEAL SEVERAL MODIFICATIONS. SPECIFICALLY, "W/O FORWARD" DENOTES THE REMOVAL OF FORWARD SCANNING, WHILE "W/O FORWARD RES" INDICATES THE EXCLUSION OF REVERSE FORWARD SCANNING, AND "W/O BACKWARD" REPRESENTS THE REMOVAL OF BACKWARD SCANNING FROM THE MSSM MODULE.

| Methods | Submap Retrieval Recall ↑ | | | | | |
|---|---|---|---|---|---|---|
| | Validation Set | | | Test Set | | |
| | K=1 | K=3 | K=5 | K=1 | K=3 | K=5 |
| W/o Forward | 0.340 | 0.602 | 0.711 | 0.335 | 0.514 | 0.609 |
| w/o Forward_rev | 0.345 | 0.600 | 0.718 | 0.340 | 0.517 | 0.604 |
| w/o Backward | 0.337 | 0.599 | 0.705 | 0.329 | 0.508 | 0.602 |
| MambaPlace (Our) | **0.352** | **0.611** | **0.721** | **0.349** | **0.530** | **0.621** |

TABLE VI

ABLATION STUDIES WITH FINE LOCALIZATION. AMONG THESE, "W/O CCAM" INDICATES THE REPLACEMENT OF CROSSMAMBA WITH FEATURE-LEVEL CROSS-MODAL FUSION FOR MODALITY INTEGRATION.

| Method | Localization Recall (Top 1) ↑ | | | | | |
|---|---|---|---|---|---|---|
| | Validation Set | | | Test Set | | |
| | k=1 | k=5 | k=10 | k=1 | k=5 | k=10 |
| Text2Loc | 0.368 | 0.571 | 0.627 | 0.330 | 0.476 | 0.521 |
| w/o CCAM | 0.391 | 0.553 | 0.608 | 0.291 | 0.442 | 0.509 |
| MambaPlace (Our) | **0.442** | **0.617** | **0.673** | **0.379** | **0.519** | **0.552** |

the overall accuracy. In the global coarse retrieval stage, our proposed MSSM demonstrated the most significant impact on the point cloud branch, improving top-1 accuracy by **65%** on the validation set and **87.6%** on the test set. This result provided strong evidence that our MSSM module, which was inspired by the human visual focusing mechanism, effectively encoded point cloud data and strengthened the positional relationships between different objects in the point cloud, leading to significant performance improvements. The ablation experiments on TAM demonstrated the effectiveness of this module, and we further verified that the mamba layer in TAM improved the model's top-1 accuracy by **3%** on the test set.

Given MSSM's substantial impact on model performance, we conducted further ablation experiments on MSSM by sequentially removing its scanning branches, as shown in Table V. Results indicated that all three scanning methods affected the model's performance. This was because the absence of any branch in the dual forward scanning prevented us from mimicking the human eye's "focusing" mechanism, while the absence of backward scanning prevented us from capturing long-range contextual information.

*2) Ablation Study with Fine Localization:* The ablation experiment conducted during the fine positioning stage validated the effectiveness of the MambaPlace method, as shown in Table VI. The implementation of CCAM enhanced performance by **5.5%** and surpassed the multimodal cascade cross-attention fusion method employed in Text2Loc, which confirmed that our approach of utilizing attention Mamba to facilitate multi-scale cross-modal fusion was viable.

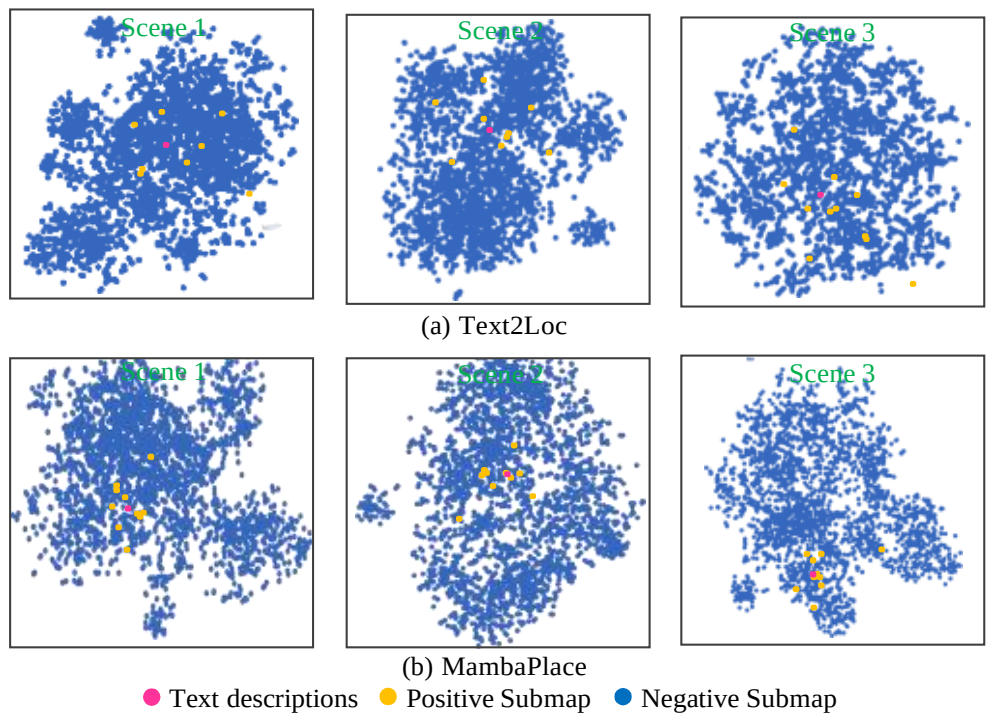

Fig. 7. Feature clustering visualization for the global place recognition based on (a) Text2Loc and (b) MambaPlace.

### *E. Embedding Space Analysis*

We utilized t-SNE to visually represent the embedding space of text-to-point cloud feature matching learned by MambaPlace, as illustrated in Fig. 7. The baseline method, Text2Loc, generated a less discriminative space, where its positive submaps were often located far from the query text descriptions or were even scattered throughout the embedding space. In contrast, MambaPlace effectively brought the positive submaps and query text representations closer together in terms of embedding distance, producing a more discriminative recognition of cross-modal spaces.

## V. CONCLUSIONS

We present MambaPlace, the first method for text-to-point-cloud place recognition framework based on the attention Mamba mechanism. We develop three distinct specialized attention Mamba modules respectively for text, point cloud, and cross-modal features. These modules are designed to strengthen dependencies in both intra- and inter-data classes. In the coarse localization stage, we introduce Text Attention Mamba and Point Cloud Multi-strategy Scanning Mamba to enhance the feature representation of both the text encoding and point cloud encoding branches. In the fine localization stage, we present Cascaded Cross-Attention Mamba to integrate the two modalities, thereby improving the performance of text-guided point cloud submap retrieval. We demonstrate that our coarse-to-fine approach can accurately localize **89%** of text queries to the query location within 15 meters when considering the top-10 retrieved locations, which surpasses the current state-of-the-art Text2Loc that achieves a localization accuracy of **86%**.